\documentclass[letterpaper, 10 pt, conference]{ieeeconf}

\usepackage{balance}

\IEEEoverridecommandlockouts

\usepackage{epsfig} 
\usepackage{times} 
\usepackage{amsmath} 
\usepackage{amssymb}  

\usepackage[binary-units=true]{siunitx}
\usepackage{algorithm}
\usepackage[noend]{algpseudocode}
\usepackage{graphicx}
\usepackage{transparent}
\usepackage{booktabs}

\usepackage[final]{microtype}
\usepackage{ellipsis}

\usepackage{centernot} 
\usepackage{multirow}

\usepackage{varwidth}

\makeatletter
\renewcommand{\ALG@beginalgorithmic}{\small}
\makeatother

\graphicspath{{graphics/}}
\def\svgwidth{3.4in}

\DeclareMathOperator*{\argmin}{arg\,min~}

\newcommand{\minthreshold}[1]{{#1_\mathrm{min}}}
\newcommand{\maxthreshold}[1]{{#1_\mathrm{max}}}
\newcommand{\discreteendtime}{N}

\newcommand{\mat}[1]{\boldsymbol{\mathbf{#1}}} 

\newcommand{\Equation}[1]{(#1)}
\newcommand{\Figure}[1]{Fig. #1}
\newcommand{\Listing}[1]{Algorithm #1}
\newcommand{\Table}[1]{Tab. #1}
\newcommand{\Section}[1]{Sec. #1}

\newcommand{\transseg}{p}
\newcommand{\vertex}{v}
\newcommand{\setofvertices}{V}
\newcommand{\belief}{b}
\newcommand{\parentmember}{\belief_p}
\newcommand{\parentbelief}{\belief.\parentmember}
\newcommand{\residentmember}{\vertex_r}
\newcommand{\residentvertex}{\belief.\residentmember}
\newcommand{\costmember}{c}

\newcommand{\setofopenbeliefs}{B_\mathrm{open}}
\newcommand{\setofclosedbeliefs}{B_\mathrm{closed}}
\newcommand{\edge}{e}
\newcommand{\setofedges}{E}
\newcommand{\terminationtime}{t_\mathrm{terminate}}
\newcommand{\runtime}{t_\mathrm{runtime}}

\newcommand{\sample}{\textproc{Sample}}
\newcommand{\nearest}{\textproc{Nearest}}
\newcommand{\connect}{\textproc{Connect}}
\newcommand{\continue}{\textproc{Continue}}
\newcommand{\pop}{\textproc{Pop}}
\newcommand{\propagate}{\textproc{Propagate}}
\newcommand{\appendbelief}{\textproc{AppendBelief}}
\newcommand{\isopen}{\textproc{IsOpen}}
\newcommand{\getsolution}{\textproc{GetDOptimalPath}}

\newcommand{\newstate}{\fullstate_\mathrm{new}}
\newcommand{\newbelief}{\belief_\mathrm{new}}
\newcommand{\nearestvertex}{\vertex_\mathrm{nearest}}

\newcommand{\newedge}{\edge_\mathrm{new}}
\newcommand{\adjacentvertex}{\vertex_\mathrm{adj}}
\newcommand{\adjacentedge}{\edge_\mathrm{adj}}

\newcommand{\startvertex}{\vertex_\mathrm{start}}
\newcommand{\goalvertex}{\vertex_\mathrm{end}}

\newcommand{\euclideandist}{d}
\newcommand{\maxspeed}{\maxthreshold{v}}

\newcommand{\budget}{C}
\newcommand{\path}{\mat{P}}

\newcommand{\zero}{\mat{0}}
\newcommand{\identity}{\mat{I}}
\newcommand{\diag}[1]{\mathrm{diag}(#1)}

\newcommand{\ddt}{\frac{d}{dt}}
\newcommand{\uncertaintysymbol}{u}
\newcommand{\uncertainty}[1]{\uncertaintysymbol ( #1  )}
\newcommand{\eigenvalue}{\lambda}
\newcommand{\cost}[1]{\textsc{Cost} ( #1 )}
\newcommand{\states}[1]{\textsc{States} ( #1 )}

\newcommand{\flighttime}{t_f}
\newcommand{\segmenttime}{{t_s}}
\newcommand{\segmenttimethreshold}{{{t}_{s,\mathrm{max}}}}

\newcommand{\sigmastate}{\mat{\Sigma}}
\newcommand{\sigmaparameter}{{\mat{\Sigma}_\Theta}}
\newcommand{\sigmainertia}{\mat{\Sigma}_{J}}
\newcommand{\sigmainertiak}[1]{\mat{\Sigma}_{J,#1}}

\newcommand{\fullstate}{\mat{x}}
\newcommand{\fullstatenominal}{{\fullstate}}
\newcommand{\fullstates}{\mathcal{X}}

\newcommand{\inputs}{\mat{u}}

\newcommand{\measurement}{\mat{z}}
\newcommand{\measurementpos}{\measurement_{p,k}}
\newcommand{\measurementatt}{\measurement_{q,k}}

\newcommand{\flatstatenominalk}[1]{\flatstate_{#1}}

\newcommand{\flatfunc}{f}

\newcommand{\systemmat}{\mat{F}_k}

\newcommand{\measurementmat}{\mat{H}_k}

\newcommand{\allocmat}{\mat{A}}

\newcommand{\processnoise}{\mat{w}}
\newcommand{\processnoisecov}{\mat{Q}}
\newcommand{\processnoisecovdiscrete}{\mat{Q}_k}

\newcommand{\forcesnoise}{\processnoise_F}
\newcommand{\momentsnoise}{\processnoise_M}

\newcommand{\forcesnoisecov}{\processnoisecov_F}

\newcommand{\momentsnoisecov}{\processnoisecov_M}

\newcommand{\measurementnoise}{\mat{v}}
\newcommand{\measurementnoisecov}{\mat{R}}
\newcommand{\measurementnoisecovdiscrete}{\measurementnoisecov_k}

\newcommand{\posnoise}{\measurementnoise_p}
\newcommand{\attnoise}{\measurementnoise_q}
\newcommand{\smallattnoise}{\mat{\Phi}}

\newcommand{\posnoisecov}{\measurementnoisecov_p}
\newcommand{\attnoisecov}{\measurementnoisecov_q}

\newcommand{\baseframe}{}
\newcommand{\worldframe}{}
\newcommand{\rotorframe}{{}}

\newcommand{\baseframehexa}{\mathcal{B}}
\newcommand{\worldframehexa}{\mathcal{W}}

\newcommand{\basehexa}{_{\baseframehexa}}

\newcommand{\base}{_{\baseframe}}
\newcommand{\world}{_{\worldframe}}

\newcommand{\zbase}{{\basehexa \mat{z}}}

\newcommand{\rotorforce}{\mat{F}_i}

\newcommand{\rotorthrust}{\mat{F}_{T,i}}

\newcommand{\rotordrag}{\mat{F}_{D,i}}

\newcommand{\rotormoment}{\mat{M}_i}

\newcommand{\gravity}{\mat{G}}
\newcommand{\gravityscalar}{g}
\newcommand{\gravityworld}{{\world \gravity}}

\newcommand{\mass}{m}

\newcommand{\motorlocation}{\mat{r}_{\baseframe\rotorframe}}
\newcommand{\motorlocationbase}{{\base\motorlocation}_i}

\newcommand{\numrotors}{k}
\newcommand{\sumrotors}{\sum_{i=1}^\numrotors}

\newcommand{\param}{\mat{\Theta}}
\newcommand{\paramdot}{\dot{\param}}
\newcommand{\numparams}{j}

\newcommand{\kthr}{c_T}

\newcommand{\kmom}{c_M}

\newcommand{\kdrag}{c_D}

\newcommand{\inertiax}{j_x}
\newcommand{\inertiay}{j_y}
\newcommand{\inertiaz}{j_z}

\newcommand{\Inertia}{\mat{J}}
\newcommand{\Inertiainv}{\Inertia^{-1}}

\newcommand{\pos}{\mat{p}}
\newcommand{\vel}{\mat{v}}

\newcommand{\att}{\mat{q}}
\newcommand{\attdot}{\dot{\mat{q}}_{IB}}
\newcommand{\attdiscrete}{\mat{q}_{k}}

\newcommand{\angvel}{\mat{\omega}}

\newcommand{\yaw}{\psi}
\newcommand{\yawdot}{\dot{\psi}}
\newcommand{\yawdotmax}{\maxthreshold{\yawddot}}
\newcommand{\yawddot}{\ddot{\psi}}

\newcommand{\posworld}{{\world \pos}}
\newcommand{\posworlddot}{{\world \dot{\pos}}}
\newcommand{\posworldddot}{{\world \ddot{\pos}}}
\newcommand{\posworlddddot}{{\world \dddot{\pos}}}
\newcommand{\posworldddddot}{{\world \ddddot{\pos}}}
\newcommand{\posworldmin}{{\world \minthreshold{\pos} }}
\newcommand{\posworldmax}{{\world \maxthreshold{\pos} }}

\newcommand{\velbase}{{\base \vel}}

\newcommand{\velbasedot}{{\base \dot{\vel}}}

\newcommand{\angvelbase}{{\base \angvel}}
\newcommand{\angvelbasemax}{\maxthreshold{\omega}}
\newcommand{\angvelbasedot}{{\base \dot{\angvel}}}

\newcommand{\rotorspeedabrv}{n}
\newcommand{\rotorspeed}{\rotorspeedabrv_i}
\newcommand{\rotorspeedsqr}{\rotorspeedabrv_i^2}

\newcommand{\rotors}{\mat{\rotorspeedabrv}}

\newcommand{\totalthrust}{T}
\newcommand{\totalthrustmin}{\minthreshold{\totalthrust}}
\newcommand{\totalthrustmax}{\maxthreshold{\totalthrust}}

\newcommand{\rot}{\mat{C}_{\worldframehexa\baseframehexa}}
\newcommand{\rotinv}{\rot^T}

\newcommand{\flatstate}{\mat{\sigma}}

\newcommand{\acc}{\mat{a}}
\newcommand{\accworld}{{\world\acc}}

\newcommand{\angacc}{\mat{\alpha}}
\newcommand{\angaccbase}{{\base\angacc}}

\usepackage{xcolor}

\definecolor{todo-red}{RGB}{200,12,12}
\definecolor{green4}{RGB}{0,128,0}

\title{\LARGE \bf
Sampling-based Motion Planning for \\Active Multirotor System Identification
}

\author{Rik B\"ahnemann, Michael Burri, Enric Galceran, Roland Siegwart, and Juan Nieto
\thanks{R. B\"ahnemann, M. Burri, R. Siegwart, and J. Nieto are with the \ac{ASL}, ETH Z\"urich, Z\"urich, Switzerland.
E. Galceran was also with the \ac{ASL} at the time this work was conducted.
   {\tt\small \{brik, burrimi, enricg, rsiegwart, jnieto\}@ethz.ch} \newline
        {This research was supported in part by the European Community's Seventh Framework Programme (grant number n.608849).}}%
}

\usepackage[printonlyused]{acronym}
\acrodef{CAD}{computer-aided design}
\acrodef{EKF}{extended Kalman filter}
\acrodef{MAV}{micro aerial vehicle}
\acrodef{PEM}{prediction error minimization}
\acrodef{UAV}{unmanned aerial vehicle}
\acrodef{RRBT}{rapidly-exploring random belief tree}
\acrodef{RIG}{rapidly-exploring information gathering}
\acrodef{CoG}{center of gravity}
\acrodef{MLE}{maximum likelihood estimate}
\acrodef{IMU}{inertial measurement unit}
\acrodef{OMPL}{open motion planning library}
\acrodef{RE}{relative error}
\acrodef{BRM}{belief roadmap}
\acrodef{AscTec}{Ascending Technology}
\acrodef{RMSE}{root mean squared error}
\acrodef{CPU}{central processing unit}
\acrodef{RAM}{random access memory}
\acrodef{ASL}{Autonomous Systems Lab}
\acrodef{ML}{maximum-likelihood}
\acrodef{RRT}{rapidly-exploring random tree} 
\acrodef{GPS}{global positioning system}
\acrodef{MPC}{model predictive control}

\begin{document}

\maketitle
\thispagestyle{empty}
\pagestyle{empty}

\begin{abstract}
This paper reports on an algorithm for planning trajectories that allow
a multirotor \ac{MAV} to quickly identify a set of unknown parameters.
In many problems like self calibration or model parameter identification some states are only observable under a specific motion.
These motions are often hard to find, especially for inexperienced users.
Therefore, we consider system model identification in an active setting,
where the vehicle autonomously decides what actions to take in order to quickly
identify the model.
Our algorithm approximates the belief dynamics of the system
around a candidate trajectory using an \ac{EKF}.
It uses sampling-based motion planning to explore the space of
possible beliefs and find a maximally informative trajectory within
a user-defined budget.
We validate our method in simulation and on a real system showing the feasibility and repeatability of the proposed approach.
Our planner creates trajectories which reduce model parameter convergence time and uncertainty by a factor of four.

\end{abstract}

\section{Introduction} \label{sec:intro}
Multirotors are becoming increasingly popular in research, industry, and consumer electronics,
with applications in aerial photography, film making, delivery, construction work, and search and rescue operations.
To achieve such complex tasks autonomously, precise maneuvering of the \ac{MAV} is required, which in turn requires accurate state estimation, planning, and control \cite{kumar2012opportunities}.
Those three components benefit greatly from an accurate aircraft model.
During aggressive maneuvers a nonlinear model can help compensate for significant aerodynamic effects that impact the vehicle \cite{mina2015}.
In manipulation or transportation, where the external forces change regularly, effective models can be used to estimate these disturbances \cite{burri2015robust}.
Furthermore, one can use a system model to create a realistic simulation environment \cite{Furrer2016},
avoiding the need to carry out expensive real-world experiments.

Estimating the parameters of such nonlinear motion models is a challenging task.
A common approach is to carry out specific experiments and simulations for each parameter,
for example measuring the drag coefficients in wind tunnel experiments, or approximating the moments of inertia with estimates from a CAD model.
These experiments, however, are expensive, time-consuming, and require significant expertise.

A more convenient alternative consists of recording sensor data during a flight and running a \ac{ML} batch optimization \cite{burri2016identification} or Kalman filtering on the data.
These approaches are especially useful if the robot needs to be re-calibrated often, for example when changing the hardware setup, or the payload in a package delivery situation.

As some parameters are only observable under special motions, \ac{ML} methods require either a short, information-rich dataset to obtain parameter estimates with a low uncertainty, or large volumes of data which increases the computational complexity of the identification process.
Designing repeatable experiments that excite all of the system's dynamics is difficult as it requires deep understanding of the model characteristics.
Transferring this knowledge into feasible teleoperated robot trajectories is a further substantial challenge.

\begin{figure}
\centering
\includegraphics[width=0.48\textwidth]{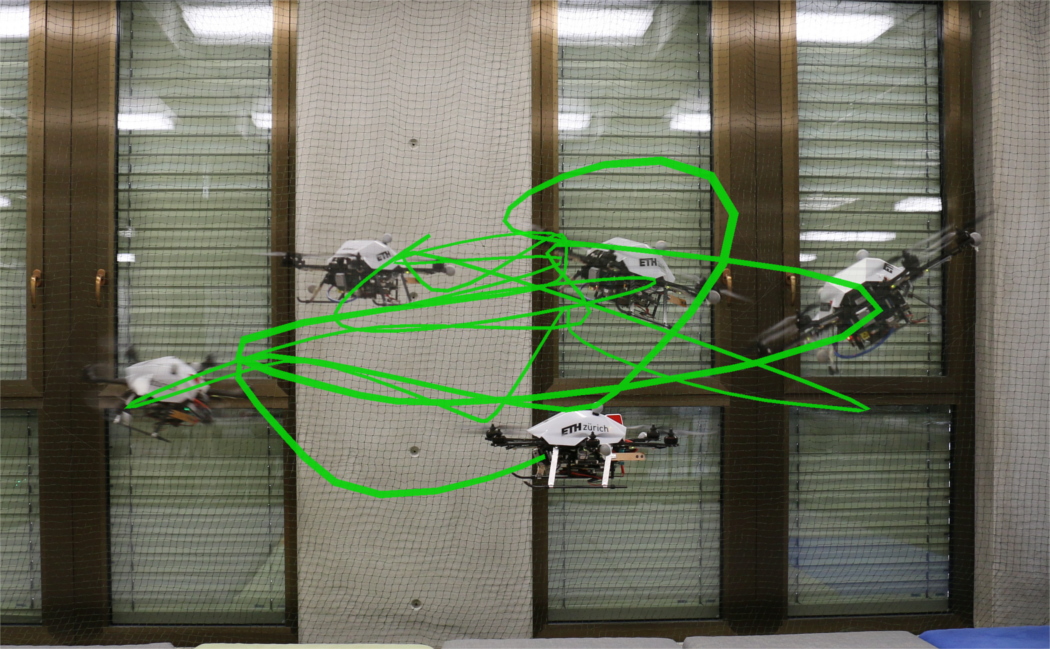}
\caption{A \ac{MAV} calibration routine generated by our proposed planner.
The trajectory is generated by maximizing the information about the system parameters,
i.e., its moments of inertia and aerodynamic coefficients.
The planner creates safe trajectories that respect state and input constraints.
The framework enables non-expert users to perform automated calibration.}
\label{fig:eyecatcher}
\vspace{-6mm}
\end{figure}

In this paper we present an automated trajectory generation framework for \ac{MAV} parameter estimation based on sampling-based motion planning under uncertainty.
The framework actively designs a repeatable and persistently informative experiment for calibration.
The algorithm performs a tree based search over various candidate trajectory segments selecting the combination that minimizes parameter uncertainty.
Our planner incorporates time, space, and control feasibility constraints.
The central contributions of this work are:
\begin{itemize}
  \item A real-time \ac{MAV} motion planning framework for parameter identification experiments which requires minimum human interaction and a small number of tuning variables.
  \item The formulation of an \ac{EKF} for parameter identification and propagation of uncertainties in \acp{RRBT}.
  \item Computational improvements to \acp{RRBT} with the inclusion of one-step propagations.
  \item The adaptation of an information criterion suited for robot system identification.
\end{itemize}

We organize the paper as follows.
Section \ref{sec:related} presents related work.
Section \ref{sec:pdef} states the formal problem definition.
We then outline the proposed information gathering algorithm in section \ref{sec:information_gathering_algorithm}.
In section \ref{sec:motion_planning} and \ref{sec:belief_propagation} we detail two main components of the algorithm: motion planning and belief propagation.
Finally, we validate the results in section \ref{sec:results} before we close the article with concluding remarks.

\section{Related Work} \label{sec:related}
Designing an optimal experiment for calibration requires making several design decisions:
which signals to measure, where to position sensors, how to filter measurements, and how to actuate the robot.
These choices should be made to maximize the observability of the modes of interest.
Finding an informative trajectory crosses both the domain of system identification and motion planning.
In order to classify a generated trajectory as being informative, one has to make an assessment of its quality for identification.

There exist a number of works in the literature that have presented methods for designing informative trajectories.
To the authors' knowledge, the work that is most similar to ours in scope is \cite{Hausman}.
In this work the authors optimize polynomial trajectories to calibrate an \acs{IMU}-\acs{GPS} \ac{MAV} model.
In contrast to our work they use a parameter information measure based on the observability Gramian and a continuous optimization.
While their approach seems promising to overcome discretization and linearization errors which occur in tree-based approaches,
they lack real-time capabilities and may be prone to local minima due to the high-dimensional non-linear space.

Another common approach to find an informative trajectory is to evaluate and optimize the statistical properties of the applied \ac{ML} estimator.
In \cite{Swevers1997} the authors generate trajectories for end effector calibration.
In a constrained non-linear optimization they generate and refine a finite sum of harmonics trajectory to perform a batch least squares estimation.
In this optimization they either minimize the regressor's condition number or maximize the Fisher information matrix.
The same optimization objective is used in \cite{Wilson2014, Wilson2015} to refine an arbitrary initial input signal for a pendulum cart identification.

Instead of optimizing a subset of trajectories, one could attempt solving the information gathering problem directly in the whole configuration space.
Solving this exactly is non-deterministic polynomial-time (NP)-hard \cite{ko1995exact}.
Consequently, we need to find a trade-off between optimality and computational feasibility.

Recent work on sampling-based information gathering has shown to rapidly lead to near optimal solutions.
Because \acp{RRT} are biased towards unexplored areas, \acp{RRT} can quickly cover a large subset of the configuration space.
This makes them real-time capable even in high dimensional spaces and due to the random nature explore different topologies to overcome local minima.

Rapidly-exploring information gathering (RIG) \acused{RIG} \cite{Hollinger2014} allows solving general information gathering problems by sampling the configuration space.
The algorithm extends previous work to submodular cost functions for which it can guarantee asymptotic optimality.

Belief graph search such as \ac{RRBT} \cite{Bry2011} or \ac{BRM} \cite{Prentice2009} on the other hand
use random trees to efficiently predict the \ac{EKF} state covariance in a graph of motions.
In \cite{achtelik2014motion} they use \ac{RRBT} to perform motion- and uncertainty-aware path planning for \acp{MAV}.

We build upon this work to formulate a real-time capable multirotor parameter estimation framework.
In contrast to \cite{achtelik2014motion} our work focuses on pure information gathering in a confined space and thus omits finding a goal path.
Furthermore, we propose various algorithmic improvements to enhance computational times of \acp{RRBT} and decrease tuning efforts.

\section{Problem Definition} \label{sec:pdef}
Our goal is to find a calibration trajectory that maximizes the information about unknown \ac{MAV} parameters.
Equivalently, we search for egomotions which will minimize our parameter estimation's uncertainty.
We formulate the problem of finding an optimal experiment design for \ac{MAV} parameter estimation
as the following optimization problem:
\begin{alignat}{2}
&\path^\ast =  &&\argmin_{\path \in \mathcal{P}} \uncertainty{\path}  \label{eq:pdef}\\
&\text{s.t.}  && \cost{\path} \leq \budget, \nonumber \\
& && \states{\path} \in \fullstates, \nonumber
\end{alignat}
where we wish to find within all possible trajectories $\mathcal{P}$, a path $\path$
which minimizes the parameter uncertainty $\uncertainty{\path}$, such that all constraints are satisfied.
The cost of the path is given by $\cost{\path}$ and is required to stay within a budget $\budget$.
The robot states along the trajectory, given by $\states{\path}$, are constrained to remaining within the set of all feasible states~$\fullstates$.

\subsection{Optimization Objective and Constraints}
\begin{figure}
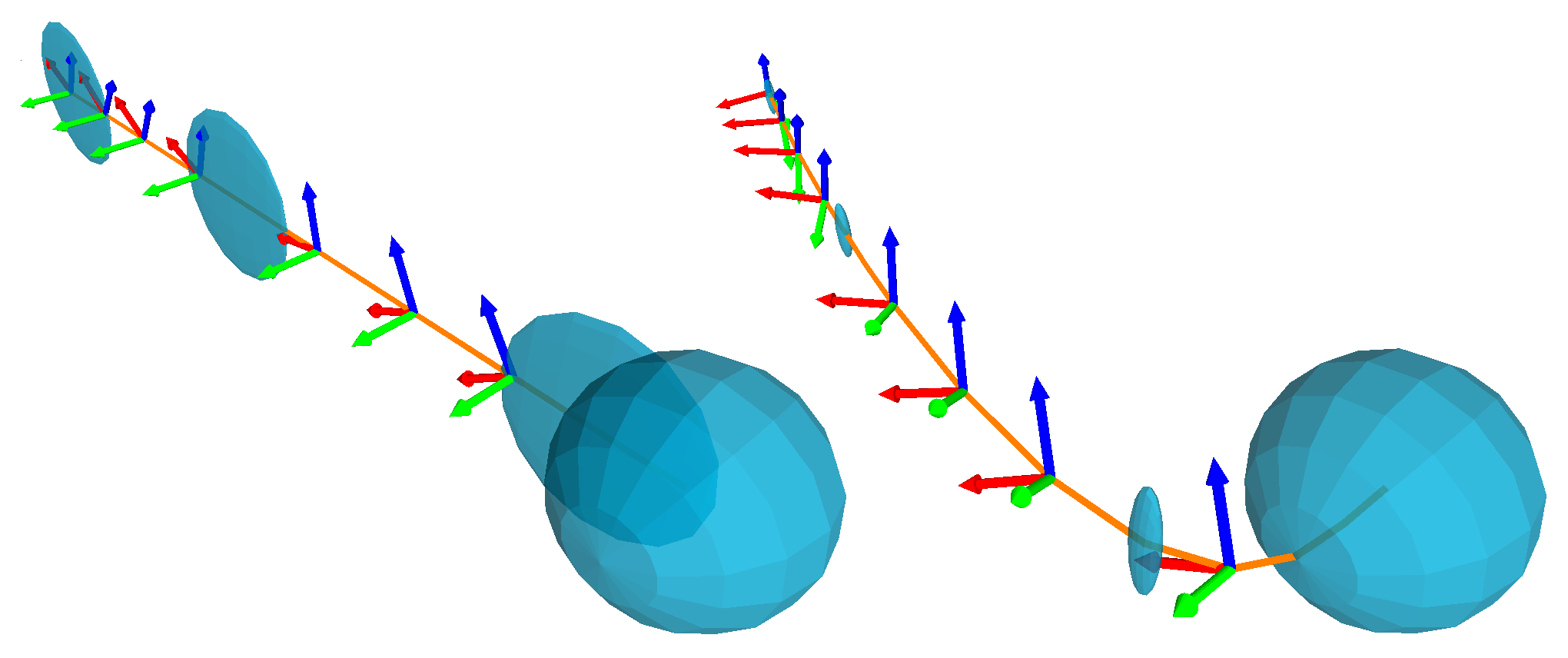
\caption{Parameter belief propagation: two different polynomial segments and the predicted moment of inertia covariance $\sigmainertia$ depicted as ellipsoids.
Left: the forward movement only gives information about the pitch inertia.
Right: the maneuver excites roll, pitch, and yaw simultaneously and shrinks the parameter covariance in all directions.
Note that the smaller the covariance matrix, the more informative the trajectory.}
\label{fig:propagation}
\vspace{-6mm}
\end{figure}

We describe the uncertainty of our unknown parameters $\param$ through the predicted
parameter covariance matrix $\sigmaparameter$ of our estimator.
A small covariance matrix indicates an informative path.
\Figure{\ref{fig:propagation}} illustrates the effect of two different planned egomotions on the parameter covariance.

The utility function $\uncertaintysymbol:\mathbb{R}^\numparams \rightarrow \mathbb{R}$ quantifies the uncertainty and
we optimize the trajectory in a D-optimal sense, i.e., we minimize $\sigmaparameter$'s determinant.
Given $\sigmaparameter$'s eigenvalues $\mat{\eigenvalue}$, the D-optimal criterion is computed as
\begin{align}
  \uncertainty{\path} = \exp( \frac{1}{\numparams} \sum_{i=1}^\numparams \log(\eigenvalue_i)  ). \label{eq:dopt}
\end{align}%

D-optimality is proportional to the volume of the $\numparams$-dimensional ellipsoid spanned by the covariance matrix.
In contrast to A-optimality or E-optimality, D-optimality weights each eigenvalue equally.
For example a decrease of one eigenvalue by $10\%$ decreases the total cost by $10\%$ independent of the eigenvalue's parameter unit.
This reduces tuning effort as the cost function does not require normalization.

We calculate the D-optimality in \Equation{\ref{eq:dopt}} in logarithmic space to avoid round off errors
in case of small eigenvalues \cite{Carrillo2012}.
Note that we choose to penalize only parameter uncertainty rather than combinations such as the
uncertainty and flight time or control effort.
This decision was made as it simultaneously simplifies the planner and reduces the difficulty of tuning the system.

To ensure that the calibration routine finishes in a given time frame, we define the trajectory's cost by its total flight time $\flighttime$.
Other measures such as power consumption could also be used for the cost measure as long as the function is strictly increasing over time.%
\begin{align}
  \cost{\path} = \flighttime.
\end{align}%

State constraints ensure the feasibility of the planned path.
Those can arise from actuator limits, sensor, and space constrains.
In particular, we place limits on absolute thrust $\totalthrust$, angular rates $\angvelbase$ in base frame, and yaw acceleration $\yawddot$ in base frame.
In addition we limit the robot to stay within a bounding box by constraining the position $\posworld$ in world frame.
\begin{gather}
           0 \leq \totalthrustmin \leq \totalthrust \leq \totalthrustmax, ~ \totalthrust = ||\posworldddot + \gravity|| , \nonumber \\
          ||\angvelbase|| \leq \angvelbasemax, ~ |\yawddot| \leq \yawdotmax, ~ \posworldmin \leq \posworld \leq \posworldmax ,\label{eq:constraints}
\end{gather}
where $\gravityworld = [0 ~ 0 ~ \gravityscalar]^T$ is the gravitational acceleration in world coordinates.
Choosing conservative limits ensures that the position controller can track the trajectory closely.
This is important since we use the nominal trajectory to propagate the beliefs in the planner.

Note that most controllers are tuned independently of the estimated system parameters.
In case the same model is used for the controller one could think of an iterative approach where the controller is updated after each identification run.

\subsection{Robot Model and \ac{EKF} Parameter Uncertainty Prediction}
We assume continuous-time nonlinear system dynamics and measurements to describe our
robot state $\fullstate$ and sensor measurements $\measurement$:
\begin{align}
  \ddt \fullstate &= \mat{f}(\fullstate, \inputs, \processnoise), && \measurement = \mat{h}(\fullstate, \measurementnoise), \label{eq:non_linear_dynamics}
\end{align}%
where $\inputs$ is the control input and
$\processnoise$ and $\measurementnoise$ are Gaussian noise accounting for
modeling and measurement errors.

We augment our state vector with constant unknown parameters $\param$
and formulate an \ac{EKF} for their estimation.
During planning we sample the polynomial trajectory and calculate the nominal states, measurements, and control inputs
along candidate trajectories.
With these we propagate the \ac{EKF} covariance to predict the probabilistic state distribution.

The parameter covariance $\sigmaparameter$ is then part of the full covariance matrix $\sigmastate$
\begin{align}
  \sigmastate &=
  \begin{bmatrix}
    \cdot && \cdot \\
    \cdot && \sigmaparameter
  \end{bmatrix} \label{eq:covariance}
\end{align}%
and is computed at time step $k$, given the prior covariance $\sigmastate_{k-1|k-1}$, as
\begin{align}
  \sigmastate_{k|k-1} &= \systemmat \sigmastate_{k-1|k-1} \systemmat^T + \processnoisecovdiscrete, \label{eq:prior}\\
  \mat{S}_k &= \measurementmat \sigmastate_{k|k-1} \measurementmat^T + \measurementnoisecovdiscrete, \\
  \mat{K}_k &= \sigmastate_{k|k-1} \measurementmat^T \mat{S}_k^{-1}, \\
  \sigmastate_{k|k} &= ( \identity - \mat{K}_k \measurementmat ) \sigmastate_{k|k-1}, \label{eq:posterior}
\end{align}%
where we obtain the discrete state transition matrix $\systemmat$, process noise covariance matrix $\processnoisecovdiscrete$,
observability matrix $\measurementmat$, and measurement noise covariance matrix $\measurementnoisecovdiscrete$ by linearizing the system about the nominal robot trajectory.
Note that we omit state prediction and updates which are unnecessary under the assumption that the \ac{MAV} is able to follow the trajectory closely.

\section{Information Gathering Algorithm} \label{sec:information_gathering_algorithm}
We approximate the solution to the minimization problem presented in \Equation{\ref{eq:pdef}} with a sampling-based information gathering algorithm.
Our planner generates a graph of random feasible motions to explore the robot's configuration space.
For every trajectory within the graph the planner computes the probability distribution over the states and parameters.
Finally, the trajectory with the least uncertainty in parameters is selected.
We follow the ideas of \ac{RIG} \cite{Hollinger2014} on sampling-based information gathering
and \ac{RRBT} \cite{Bry2011} on belief space search.

In \ac{RIG} the planner searches for a maximum informative path.
It samples random motions and evaluates an arbitrary information criterion for each candidate path.
In our case the covariance measure stated in \Equation{\ref{eq:dopt}} describes the information we gather about our parameter estimates.
The parameter uncertainty, however, is a function of the state covariance $\sigmastate$.
This requires us to keep track of the state's probabilistic distribution over all sampled trajectories.
\acp{RRBT} do not only explore the configuration space, but calculate the probabilistic distribution for each state by propagating an \ac{EKF},
effectively searching the robot's belief space.

We adapt the original \ac{RRBT} algorithm such that it does not search for a goal connection but rather, only searches the belief space and minimizes the parameter uncertainty.
We also introduce the budget constraint $\budget$.
The planner differentiates between \textit{open} and \textit{closed} trajectories.
Furthermore, we do not consider deviations of the robot from the planned path, but assume that the controller can follow the trajectory closely.
We also interconnect all vertices in the graph rather than only connecting neighbors as in \ac{RRBT}.

\Listing{\ref{lst:algorithm}} shows the graph search.
It consists of two main parts: (i) building a bidirectional graph of motions
and (ii) searching the graph for the trajectory that minimizes parameter uncertainty.
\vspace{-2mm}
\begin{algorithm}
\caption{Motion planning for \ac{MAV} parameter estimation} \label{lst:algorithm}
\begin{algorithmic}[1]
\State $\belief.\sigmastate := \sigmastate_0$; $\belief.\costmember := 0$; $\parentbelief:=\mathrm{NULL}$; $\residentvertex := \vertex$
\State $\vertex.\fullstate := \fullstate_0$; $\vertex.\setofopenbeliefs := \{\belief\}$; $\vertex.\setofclosedbeliefs := \{\}$;
\State $\setofvertices := \{\vertex\}$; $\setofedges:= \{\}$; $Q := \{\}$
\While { $\runtime < \terminationtime$ }
  \State \% {Phase 1: motion planning}
  \State $\newstate :=  \sample() $
  \State $\nearestvertex := \nearest(\setofvertices, \newstate)$
    \If {$!\exists (\newedge := \connect ( \newstate$, $\nearestvertex.\fullstate ))$ }
      \State \continue
    \EndIf
  \State $E := E \cup \newedge$
  \State $V := V \cup \vertex(\newstate)$
  \State $Q := Q \cup \nearestvertex.\setofopenbeliefs$
  \State $E := E \cup \connect(\nearestvertex.\fullstate, \newstate)$
  \ForAll { $\vertex \in \setofvertices$ }
    \If {$\exists \newedge := \connect(\vertex.\fullstate, \newstate)$}
      \State $E := E \cup \newedge$
      \State $Q := Q \cup \vertex.\setofopenbeliefs$
    \EndIf
    \State $E := E \cup \connect(\newstate, \vertex.\fullstate)$
  \EndFor
  \State \% {Phase 2: belief propagation}
  \While {$Q \neq \emptyset$}
  \State $\belief := \pop(Q)$
    \ForAll {$\adjacentvertex(\belief)$} \Comment{reachable vertices}
      \State $\newbelief := \propagate(\adjacentedge, \belief)$
      \If {$\appendbelief(\adjacentvertex, \newbelief)$}
        \State $Q := Q \cup \newbelief$
        \If {$\isopen(\newbelief)$}
          \State $\adjacentvertex.\setofopenbeliefs = \adjacentvertex.\setofopenbeliefs \cup \newbelief$
        \Else
          \State $\adjacentvertex.\setofclosedbeliefs = \adjacentvertex.\setofclosedbeliefs \cup \newbelief$
        \EndIf
      \EndIf
    \EndFor
  \EndWhile
\EndWhile
\State \Return $\path := \getsolution()$
\end{algorithmic}
\end{algorithm}

\vspace{-2mm}

The motion graph consists of a set of vertices $\vertex \in \setofvertices$ which represent robot states $\fullstate$.
These vertices are connected by bidirectional edges $\edge \in \setofedges$ allowing the robot to transition between two states.
There are multiple paths through the graph to reach one vertex.
This history of propagation is stored in belief nodes $\belief$ which reside at each vertex.
Each belief node describes a unique trajectory through the graph.
It contains the state covariance $\sigmastate$, the cost $\costmember$,
its parent belief node $\parentmember$, and resides at vertex $\residentmember$.

In the motion planning phase (i) the planner uniformly samples feasible states and attempts to connect them to all other existing vertices: forwards and backwards.
In the belief propagation phase (ii) the planner evaluates all possible new connections.
It propagates all adjacent \textit{open} beliefs of a new vertex through every new path combinations.
The planner iterates between the two phases until its runtime $\runtime$ exceeds the termination time $\terminationtime$.

\section{Motion Planning} \label{sec:motion_planning}
In our motion planning, the motions are represented as $4D$-polynomial segments in position, yaw, and time as in \cite{achtelik2014motion}.
This helps to reduce the high dimensional state space of the \ac{MAV} and quickly cover large sections of the configuration space with a low number of samples.
With this approach a first feasible solution is found after a short time and from then on the solution improves as the planner has more time to explore the space.
Due to the differential flatness property of \acp{MAV} it is guaranteed that the \ac{MAV} is able to follow them, as long as the dynamic constraints \Equation{\ref{eq:constraints}} are not violated.

In our case the \ac{MAV} flat state $\flatstate$ contains position, yaw, and its derivatives \cite{mellinger2011minimum}:
\begin{align}
\flatstate = \begin{bmatrix}
\posworld^T ~ \posworlddot^T ~ \posworldddot^T ~ \posworlddddot^T ~ \posworldddddot^T ~ \yaw ~ \yawdot ~ \yawddot
\end{bmatrix}^T .  \label{eq:flat_state}
\end{align}%

The algorithm starts with $\sample$ which returns a feasible position and yaw angle, from a uniform distribution, leaving the derivatives free.
The first $\connect$ creates a minimum-snap and minimum-angular-velocity $4D$-polynomial segments towards that state.
This connection locks all state derivatives.
All other connections are then fully constrained up to snap and yaw acceleration.
For translational \mbox{$P^{\text{th}}$-order} minimum-snap polynomials $\transseg(t)$ with coefficients $\mat{a} = [a_0, a_1, \ldots, a_N]^T$ the optimization is given by:
\begin{gather}
  \transseg(t) = \mat{t} \cdot \mat{a}, ~\mat{t} = \begin{bmatrix} 1 & t & t^2 & \cdots & t^{P-1} \end{bmatrix}, \\
  \min \int_{0}^\segmenttime \left|\left| \frac{d^4 \transseg(t)}{dt^4} \right|\right| dt , \nonumber \\
  \mathrm{s.t.}   \frac{d^m \transseg(0)}{dt^m} = \frac{d^m(\startvertex.\fullstate.\posworld)}{dt^m}, ~ m=0,\ldots,4 , \nonumber \\
  \transseg(\segmenttime) = \goalvertex.\fullstate.\posworld , \nonumber \\
 \frac{d^m \transseg(\segmenttime)}{dt^m} = \frac{d^m \goalvertex.\fullstate.\posworld}{dt^m} \text{~or free}, ~ m=1,\ldots,4.
 \label{eq:min_snap_optimization}
\end{gather}

Minimum-angular-velocity yaw segments can be derived analogously.
Note that the trajectories could also be optimized with respect to other derivatives which is subject of future investigations.
We solve the minimization \Equation{\ref{eq:min_snap_optimization}} with the optimization presented in \cite{richter2013polynomial}.

We enforce continuity up to snap and yaw acceleration at the start and goal state
and thus generate \mbox{$10^{\text{th}}$-order} polynomial segments in position and \mbox{$6^{\text{th}}$-order} in yaw.
This implies continuous rotor speeds at the vertices as shown in section \ref{sec:ml_measurements}.
It allows smooth state and belief propagation as it prevents further nonlinearities due to discontinuous inputs.

Furthermore, we sample the segment time $\segmenttime$ for every new segment uniformly.
Given the Euclidean distance $\euclideandist$, maximum allowed absolute velocity $\maxspeed$, and threshold $\segmenttimethreshold$ on the maximum segment time
the sample range is:
\begin{align}
\segmenttime \in \left[ \frac{\euclideandist}{\maxspeed}, \min( \budget - \min(\cost{\startvertex.\setofopenbeliefs}), \segmenttimethreshold  ) \right].
\end{align}%

The lower limit is the physically achievable minimum segment time.
The upper limit is either the maximum budget left at the start vertex or a fixed maximum allowed time.
The budget constraint ensures that at least one belief can be propagated.
The time threshold constraint reduces the search space further, i.e., we do not allow slow segments which have shown to be uninformative.

We apply the recursive algorithm presented in \cite{muellercomputationally} to check a segment for feasibility as stated in \Equation{\ref{eq:constraints}}.
We extend this test for maximum yaw rates and accelerations by evaluating the roots of the appropriate polynomial.
Note that these checks are independent of the estimated model parameters and neglect drag force.

\section{Belief Propagation} \label{sec:belief_propagation}
\begin{figure}
\centering
\tiny{
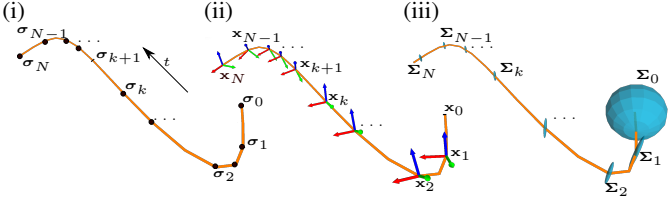
}
\caption{The three steps of belief prediction along a candidate polynomial segment:
(i) sample flat states,
(ii) recover linearization states,
(iii) propagate the covariance along the segment using the \ac{EKF}.}
\label{fig:propagation_viz}
\vspace{-6mm}
\end{figure}

Given a candidate $4D$-polynomial segment and a \ac{MAV} motion and measurement model,
we can predict the change of the state covariance from an initial belief $\sigmastate_{0|0}$ to a terminate belief $\sigmastate_{\discreteendtime|\discreteendtime}$.
The $\propagate$ method shown in \Listing{\ref{lst:algorithm}} has three steps:
(i) sample the nominal flat states along the segment,
(ii) recover and linearize the system about the full expected states, measurements, and inputs, and
(iii) propagate the covariance using the \ac{EKF}.
We visualize the process in Fig. \ref{fig:propagation_viz} and describe the key components in the following.
\def\svgwidth{3.5in}
\def\svgwidth{3.5in}

\subsection{\ac{MAV} Model} \label{sec:model}
We formulate the \ac{EKF} for a simple multirotor dynamics model as proposed in \cite{mahony2012multirotor}.
Our approach allows exchanging this model for any model that needs to be calibrated, e.g., a visual-inertial odometry model \cite{weiss2012real}.
In this particular model, the dynamic couplings of the parameters make it hard to generate manual calibration routines \cite{burri2016identification}.

The \ac{MAV} state in this model consists of position $\posworld$ in world frame, velocity $\velbase$ in base frame,
attitude quaternion $\att$ describing the rotation from inertial to base frame,
angular velocity $\angvelbase$ in base frame and the desired physical parameters.
Aerodynamic effects are described through three coefficients for thrust $\kthr$, drag $\kdrag$, and moment $\kmom$.
The moment of inertia matrix is \mbox{$\Inertia = \diag{[\inertiax ~ \inertiay ~ \inertiaz]}$.}
We define the input as the collection of the $\numrotors$ rotor speeds $\rotorspeed, ~i\in \{1 \dots \numrotors\}$.
\begin{align}
  \fullstate &= \begin{bmatrix} \posworld^T & \velbase^T & \att^T & \angvelbase^T & \param^T \end{bmatrix}^T, \label{eq:states}\\
  \param &= \begin{bmatrix} \kthr & \kdrag & \kmom & \inertiax & \inertiay & \inertiaz \end{bmatrix}^T, \label{eq:params} \\
  \inputs &= \rotors = [\rotorspeedabrv_1 ~ \ldots ~ \rotorspeedabrv_\numrotors]^T . \label{eq:inputs}
\end{align}%

The forces and moments acting on the \ac{MAV} are the forces $\rotorforce$ and moments $\rotormoment$ acting on each rotor as well as the gravitational force such that the dynamics are given by
\begin{align}
  \posworlddot &= \rot(\att) \cdot \velbase \\
  \velbasedot &= \frac{1}{\mass} \sumrotors \rotorforce - \angvelbase \times \velbase - \rotinv(\att) \cdot \gravityworld, \label{eq:newton}\\
  \attdot &= \frac{1}{2} \begin{bmatrix} 0 \\ \angvelbase \end{bmatrix} \otimes \att , \\
  \angvelbasedot &= \Inertiainv \left( \sumrotors \left( \rotormoment + \motorlocationbase \times \rotorforce  \right) - \angvelbase \times \Inertia \angvelbase \right), \label{eq:euler} \\
  \paramdot &= \zero,
\end{align}%
where $\rot$ describes the rotation matrix from base to world frame,
$\mass$ is the mass,
$\otimes$ is the quaternion product, and $\motorlocationbase$ is the rotor location in the base frame with respect to the \ac{CoG}.
The rotor force $\rotorforce$ is the sum of all rotor thrusts $\rotorthrust$ which points upwards into $\zbase$--direction in the base frame.
The combined rotor drag and flapping force $\rotordrag$ at rotor $i$ counteracts the \ac{MAV}'s body velocity.
The rotor moment $\rotormoment$ is the yaw moment induced by \mbox{rotor speed $\rotorspeed$}.
\begin{align}
\rotorthrust &= \kthr \rotorspeedsqr \zbase, \\
\rotordrag &= - \kthr \rotorspeedsqr \begin{bmatrix}
\kdrag & 0 & 0 \\
0 & \kdrag & 0 \\
0 & 0 & 0
\end{bmatrix} \left( \velbase + \angvelbase \times \motorlocationbase \right), \label{eq:drag}\\
\rotorforce &= \rotorthrust + \rotordrag + \forcesnoise, ~\forcesnoise \sim \mathcal{N}(\zero, \forcesnoisecov),  \\
\rotormoment &= -\epsilon_i \kmom \rotorspeedsqr \zbase + \momentsnoise, ~\momentsnoise \sim \mathcal{N}(\zero, \momentsnoisecov),
\end{align}%
where $\forcesnoise$ and $\momentsnoise$ are normally distributed white noise with variance $\forcesnoisecov$ and $\momentsnoisecov$ to account for modeling errors.


The \ac{MAV} is provided with discrete position $\measurementpos$ and attitude measurements $\measurementatt$.
In this work these measurements were provided by an external motion capture system.
With this set of measurements the \ac{EKF} is able to refine the covariances.
The measurement model is given by
\begin{align}
\measurementpos &= \posworld_k + \posnoise, ~ &\posnoise \sim \mathcal{N}(\zero, \posnoisecov), \\
\measurementatt &= \attdiscrete \otimes \attnoise, \\
\attnoise &= \begin{bmatrix}
1 & \frac{1}{2} \smallattnoise
\end{bmatrix}^T , &\smallattnoise \sim \mathcal{N}( \zero, \attnoisecov), \\
\measurement_k &= [ \measurementpos^T ~ \measurementatt^T ]^T,
\end{align}%
where $\posnoise$ and  $\attnoise$ is normally distributed white noise with variance $\posnoisecov$ and $\attnoisecov$ accounting for small measurement errors in position or attitude.

\subsection{Recovering the Nominal States and Inputs} \label{sec:ml_measurements}
In our formulation we assume that the \ac{MAV} follows the planned trajectories exactly.
Hence evaluating position, yaw, and its derivatives along a segment corresponds with evaluating the nominal states.
Following \cite{mellinger2011minimum} there exists a function $\flatfunc$ which maps the flat state $\flatstate$ into the full state $\fullstate$:
\begin{align}
  \flatfunc: \flatstate \rightarrow \fullstate.
\end{align}%
Additionally, one can compute the rotor speeds given the constant allocation matrix $\allocmat$, which maps rotor speeds into torque and thrust:
\begin{align}
\begin{bmatrix}
 \Inertia \angaccbase + \angvelbase \times \Inertia \angvelbase \\
 m |\accworld + \gravityworld|
\end{bmatrix} &= \allocmat \cdot \rotors^2 \label{eq:allocation},
\end{align}%
where $\angaccbase$ is the angular acceleration in the base frame.
This can also be recovered from the flat state.

\subsection{One-Step Propagation}
For each time step $k=1,\ldots,\discreteendtime$ the algorithm linearizes the system equations about the maximum likelihood state.
We follow the approach presented in \cite{weiss2012vision} and \cite{Trawny2005} for linearizing the attitude dynamics.
The \ac{EKF} covariance propagation, however, is computationally expensive and has to be repeated many times for the same edge during the planners propagation phase, e.g., if the motion generation closes a circle.
In \cite{Prentice2009} the authors introduce a method of factoring the covariance to
collapse this step by step calculation in \Equation{\ref{eq:prior} -- \ref{eq:posterior}} into a single linear transfer function $\mathcal{S}_{0:\discreteendtime}$:
\begin{align}
\begin{bmatrix}
\cdot & \sigmastate_{\discreteendtime|\discreteendtime} \\
\cdot & \cdot
\end{bmatrix} &=
\begin{bmatrix}
\identity & \sigmastate_{0|0} \\
\zero & \identity
\end{bmatrix}
\star \mathcal{S}_{0:\discreteendtime}, \label{eq:one_step}
\end{align}%
where $\star$ resembles the Redheffer star product for interconnecting two systems.
We precompute and store $\mathcal{S}_{0:\discreteendtime}$ for every new edge:
\begin{align}
\mathcal{S}_{0:\discreteendtime} &= \mathcal{S}_0 \star \mathcal{S}_1 \star \cdots \star \mathcal{S}_\discreteendtime, &\mathcal{S}_k &= \mathcal{S}_k^C \star \mathcal{S}_k^M .
\end{align}%
The motion update $\mathcal{S}_k^C$ and measurement update $\mathcal{S}_k^M$ are
\begin{align}
\mathcal{S}_k^C &=
\begin{bmatrix}
\systemmat & \processnoisecovdiscrete \\
\zero & \systemmat^T
\end{bmatrix},
&
\mathcal{S}_k^M &=
\begin{bmatrix}
\identity & \zero \\
-\measurementmat^T \measurementnoisecovdiscrete^{-1} \measurementmat & \identity
\end{bmatrix}.
\end{align}%

\section{Pruning and Optimality}
Even though we are speeding up the calculation with polynomial segments and one-step propagations,
the algorithm still has to evaluate every new possible path combination every time a new vertex is added to the graph.
This is a combinatorial problem, leading to an increasing growth of beliefs and thus propagation time.
In practice, the planner soon starts spending most of its time in the exhaustive search over all possible vertex connections.

Many of the evaluated beliefs however, are uninformative.
For example if its trajectory is slow or does not excite desired modes.
$\appendbelief$ prunes those beliefs, trading off optimality guarantees for computational feasibility.

We consider belief $\belief_a$ better than belief $\belief_b$ if it has smaller cost $\costmember$, state uncertainty  $\sigmastate$, and parameter uncertainty $\sigmaparameter$.
We add the full state covariance $\sigmastate$ to the comparison, because a converged state can have a positive influence on a parameter update.
We compare two matrices in a D-optimal way.
\begin{align}
\belief_a < \belief_b
\Leftrightarrow  { \belief_a.\costmember < \belief_b.\costmember } \wedge
{ \belief_a.\sigmastate < \belief_b.\sigmastate } \wedge
{ \belief_a.\sigmaparameter < \belief_b.\sigmaparameter }   \label{eq:compare}
\end{align}%

\begin{figure}
\centering
\includegraphics{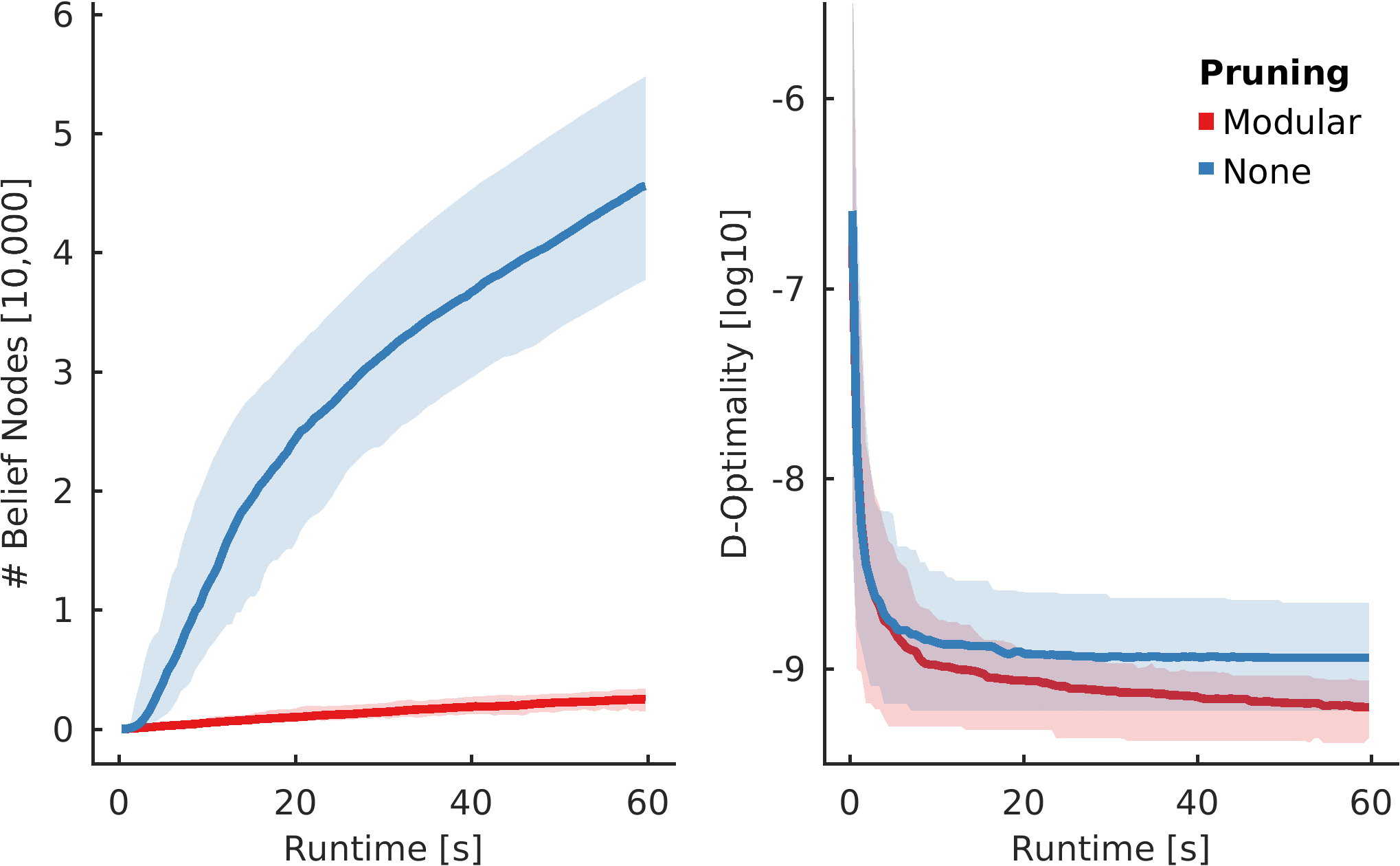}
\caption{Median and $95\%$-percentile of the number of belief nodes and the corresponding minimum uncertainty vs. the planner runtime in $100$ runs.
Modular pruning speeds up the convergence rate and leads to smaller uncertainties by pruning uninformative beliefs nodes.
Further the sampling-based planner finds an informative trajectory in less than a minute which makes it significantly faster than existing optimization-based approaches \cite{Hausman}.
The benchmark was created with \cite{omplbenchmark}.}
\label{fig:pruning}
\vspace{-6mm}
\end{figure}

\Figure{\ref{fig:pruning}} shows that this pruning leads to a fraction of beliefs and smaller uncertainty in less time.
The pruning shortens the propagation queue and allows quicker exploration of the configuration space.
The drawback of this approach is that we lose optimality guarantees as shown for \ac{RIG} by \cite{Hollinger2014}.
Due to coupled terms in the state covariance, we have a submodular utility function \Equation{\ref{eq:dopt}} but apply modular pruning \Equation{\ref{eq:compare}}.

Note that the planner was not globally optimal in the first place, because we
sample only a subspace of the configuration space;
however, as will be shown in our results, the algorithm quickly finds a solution which sufficiently excites all modes.

%
%

\section{Results} \label{sec:results}
We implemented our planner in the \ac{OMPL} \cite{ompl} using C++ and evaluated it in simulations and real experiments.
The algorithm is tested on an Intel Core i$7$-$5600$U \acs{CPU} running at \SI{2.60}{\giga\hertz} with \SI{12}{\giga\byte} of \acs{RAM}.

\subsection{Simulation Based Experiments}
Simulation based experiments were first conducted to allow the extensive comparison of our method against ground truth.
We use RotorS \cite{Furrer2016} to simulate an \ac{AscTec} Firefly hexacopter with the model given in \Section{\ref{sec:model}}.
An artificial Vicon motion tracking system provides noisy position and attitude measurements
with standard deviation \SI{0.5}{\milli\metre} and \SI{0.1}{\degree}.

\begin{figure}
\centering
\includegraphics[width=0.48\textwidth]{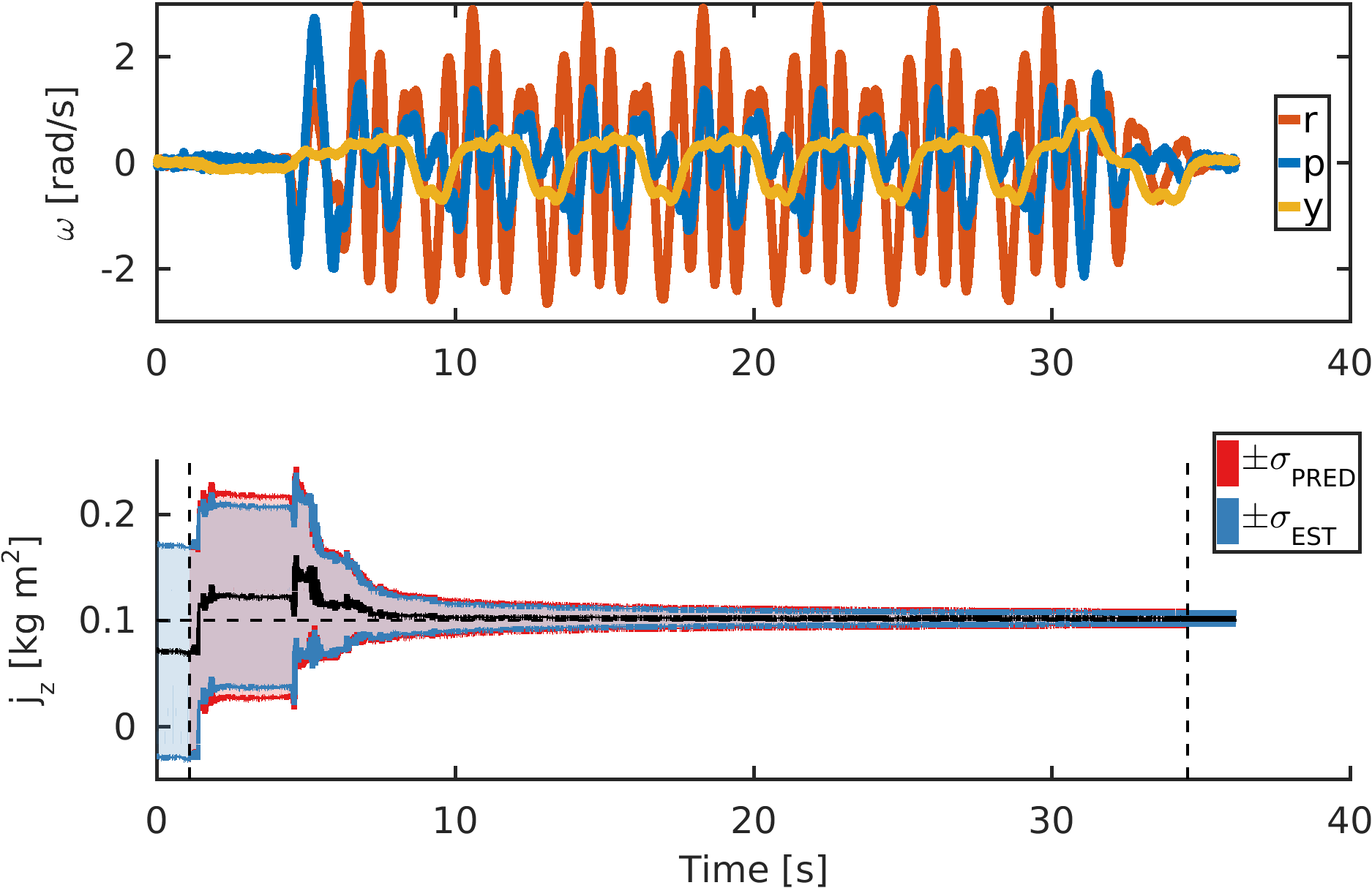}
\caption{A typical simulated experiment: (top) angular rates vs. flight time, (bottom) \ac{EKF} parameter estimation with predicted and actual sigma bounds vs. flight time.
The planner designs an aggressive trajectory that excites all directions.
It closely predicts the estimator's covariance.
This allows the algorithm to generate a trajectory with low parameter uncertainty.
Even a difficult to observe parameter like $\inertiaz$ converges within less than $\SI{7}{\second}$ of flight data.}
\label{fig:est_j_z}
\end{figure}

\Figure{\ref{fig:est_j_z}} shows the angular rates of a typical trajectory with corresponding \ac{EKF} parameter
estimation and one sigma bounds.
Typically, the trajectories are as aggressive as possible given the input feasibility constraints.
This maximizes the signal to noise ratio and the number of informative maneuvers given the budget constraint.
It also tends to repeat information rich segments.
Furthermore, the trajectory excites all directions to reduce uncertainty in all parameters.
The planner demonstrates a tendency to select segments that excite roll, pitch, and yaw simultaneously to enable the observer to distinguish
between the strongly coupled parameters $\inertiaz$ and $\kmom$ (see \Equation{\ref{eq:euler}}).

In order for our approach to operate effectively, the planner needs the ability to predict the covariance of a segment.
This can be seen in the lower part of \Figure{\ref{fig:est_j_z}} where the planner's predicted sigma bounds and the estimator's sigma bounds
illustrate the quality of the belief prediction.

Next we validate the repeatability of our approach.
We plan \num{100} parameter estimation experiments offline, simulate the flights, and run the proposed \ac{EKF} offline to estimate the parameters.
At the beginning of each experiment we sample the unknown physical parameters $\pm50\%$ around the ground truth values
to set the planner model and initialize the actual \ac{EKF} estimation.
The planner searches for only \SI{30}{\second} and generates trajectories with a flight time of \SI{30}{\second}.

We compare our results to random trajectory calibrations.
These random trajectories are generated with the same graph based motion planning.
A trajectory with a maximum number of different segments is then picked.
This heuristic biases the random planner towards exciting all directions and fast maneuvers.

\begin{figure}
\centering
\includegraphics[width=0.48\textwidth]{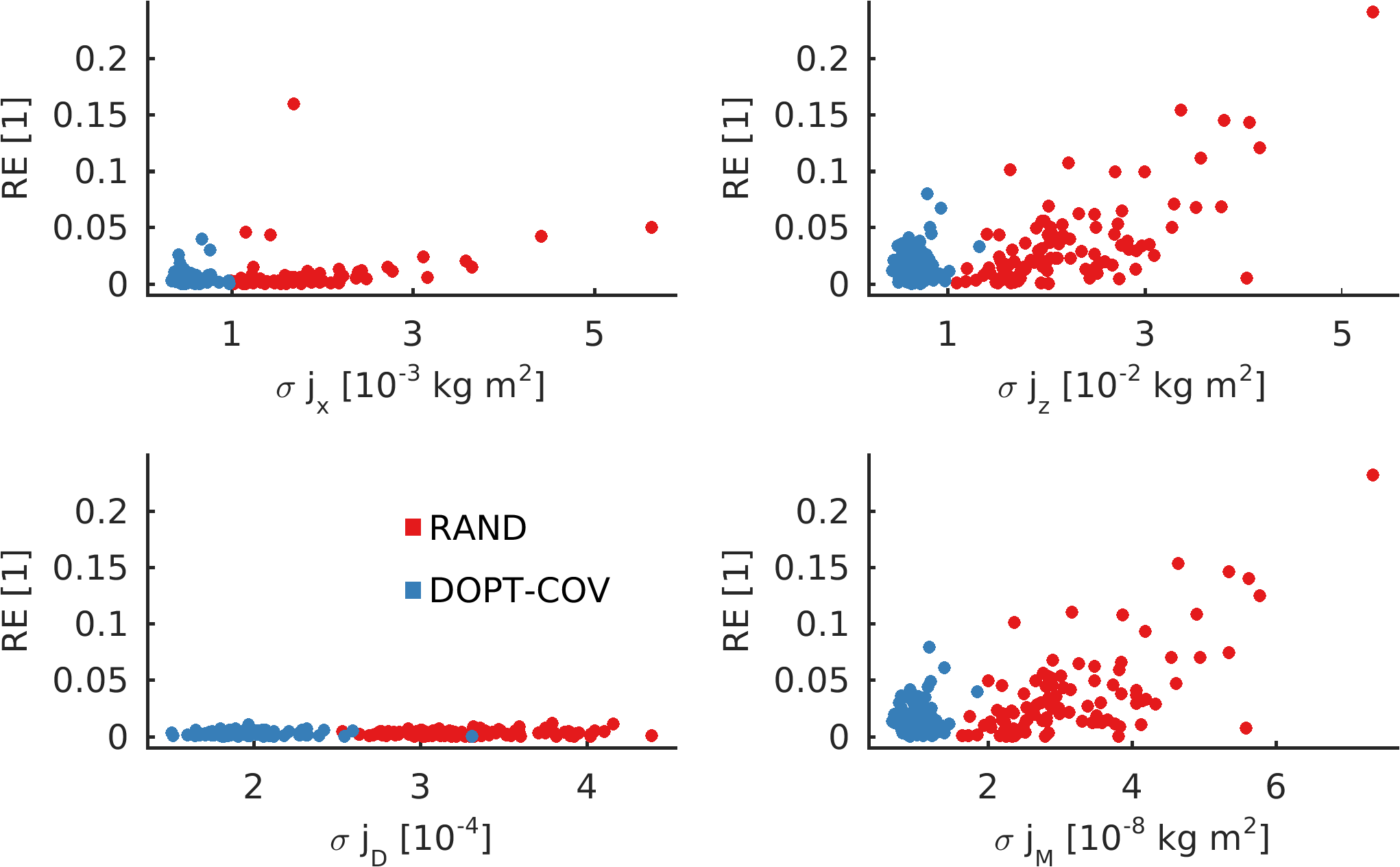}
\caption{Relative final estimation error (RE) vs. final sigma bounds for \num{100} simulated parameter estimation experiments with random initial parameter guess.
The plot shows the results for random (RAND) and optimized (DOPT-COV) calibration trajectories.
Our planner leads to smaller uncertainties and errors in parameter estimates.
In particular, the strongly coupled parameters $\inertiaz$ and $\kmom$ are estimated with more confidence.}
\label{fig:stat_sigma}
\vspace{-6mm}
\end{figure}

\Figure{\ref{fig:stat_sigma}} shows the final sigma bounds and estimation errors for our optimized and random trajectories.
Utilization of the proposed planner leads to smaller final uncertainty for all parameters.
This is of particular importance for the coupled parameters that are inherently more challenging to accurately identify.
We do not show $\inertiay$ since the platform is almost symmetric.
The thrust constant $\kthr$ converges already at hovering without any special calibration routine.

\begin{figure}
\centering
\includegraphics[width=0.48\textwidth]{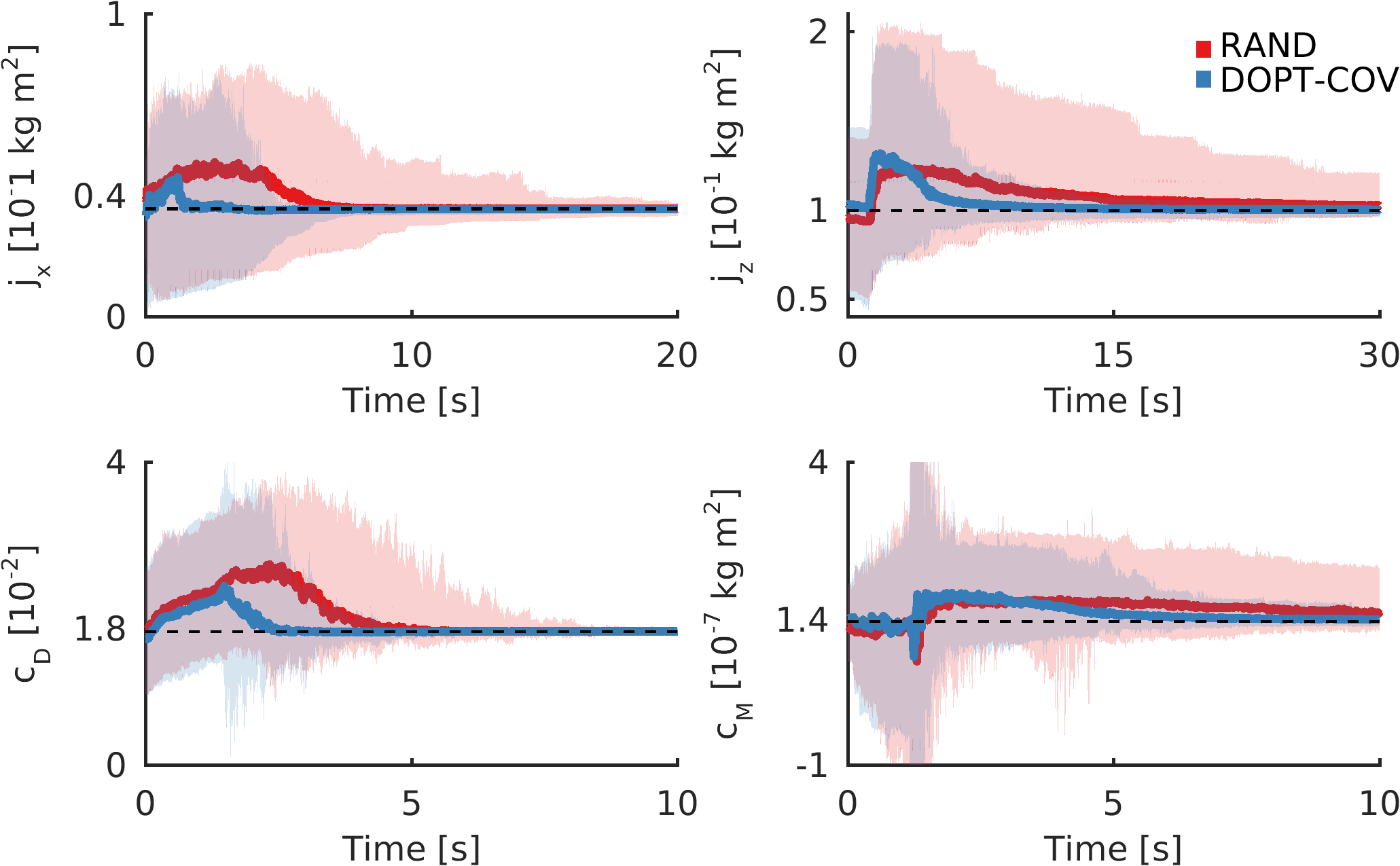}
\caption{Median and $95\%$-percentiles of \ac{EKF} parameter estimations vs. flight time for \num{100} simulated parameter estimation experiments with random initial guess.
The plot shows the results for random (RAND) and optimized (DOPT-COV) calibration trajectories.
Our planner reduces convergence time and converges in all experiments showing repeatability of the approach.}
\label{fig:stat_summary}
\vspace{-6mm}
\end{figure}

Finally, we show the convergence speed and repeatability with our approach in the statistical summary in \Figure{\ref{fig:stat_summary}}.
\Table{\ref{tab:stat_summary}} quantifies the results of the statistical evaluation.
We achieve more than $4\times$ faster convergence for $\inertiax$ with our optimized trajectories.
On average, the planner reduces the amount of data the estimator requires by $\SI{13}{\second}$ or $\SI{70}{\percent}$ compared to random trajectories.
Furthermore, the median final uncertainty \Equation{\ref{eq:dopt}} is more than $4\times$ smaller with our proposed planner which makes our estimates more confident.
\begin{table}[b]
\begin{center}
\begin{tabular}{lcrr}
  \toprule
                                            & PARAMETER   & \multicolumn{2}{c}{PLANNER}               \\
                                            &             & DOPT-COV            & RAND                \\
  \midrule
  \multirow{6}{*}{CONV TIME}                & $\kthr$     & \SI{0.05}{\second}  & \SI{0.01}{\second}  \\
                                            & $\kdrag$    & \SI{2.32}{\second}  & \SI{4.37}{\second}  \\
                                            & $\kmom$     & \SI{6.64}{\second}  & \SI{22.06}{\second} \\
                                            & $\inertiax$ & \SI{1.62}{\second}  & \SI{6.49}{\second}  \\
                                            & $\inertiay$ & \SI{3.17}{\second}  & \SI{5.69}{\second}  \\
                                            & $\inertiaz$ & \SI{6.46}{\second}  & \SI{20.01}{\second} \\
  DOPT                                      & ALL         & \num{2.72e-10}      & \num{1.21e-09}      \\
  \bottomrule
\end{tabular}
\end{center}
\caption{}{Convergence time of the median of the estimates (CONV TIME) and the median of the final D-optimal uncertainty (DOPT) from our $100$ simulated estimates.
We consider the median being converged when it reaches a ground truth error of less than $5\%$ and stays constant.
Calibrations on our optimized datasets require on average $\SI{70}{\percent}$ less data and are $4\times$ more confident than calibrations on randomly generated data.}
\label{tab:stat_summary}
\end{table}

The results support our major claims:
(i) the algorithm closely predicts the \ac{EKF} parameter covariance,
(ii) it covers a sufficient subspace of motions,
(iii) it repeatably and quickly generates persistently exciting trajectories,
(iv) it outperforms random trajectories, and
(v) unlike random trajectories it actively selects trajectories that excite modes that are hard to observe.

\subsection{Real Platform Experiments}
\begin{figure}
\centering
\includegraphics[width=0.48\textwidth]{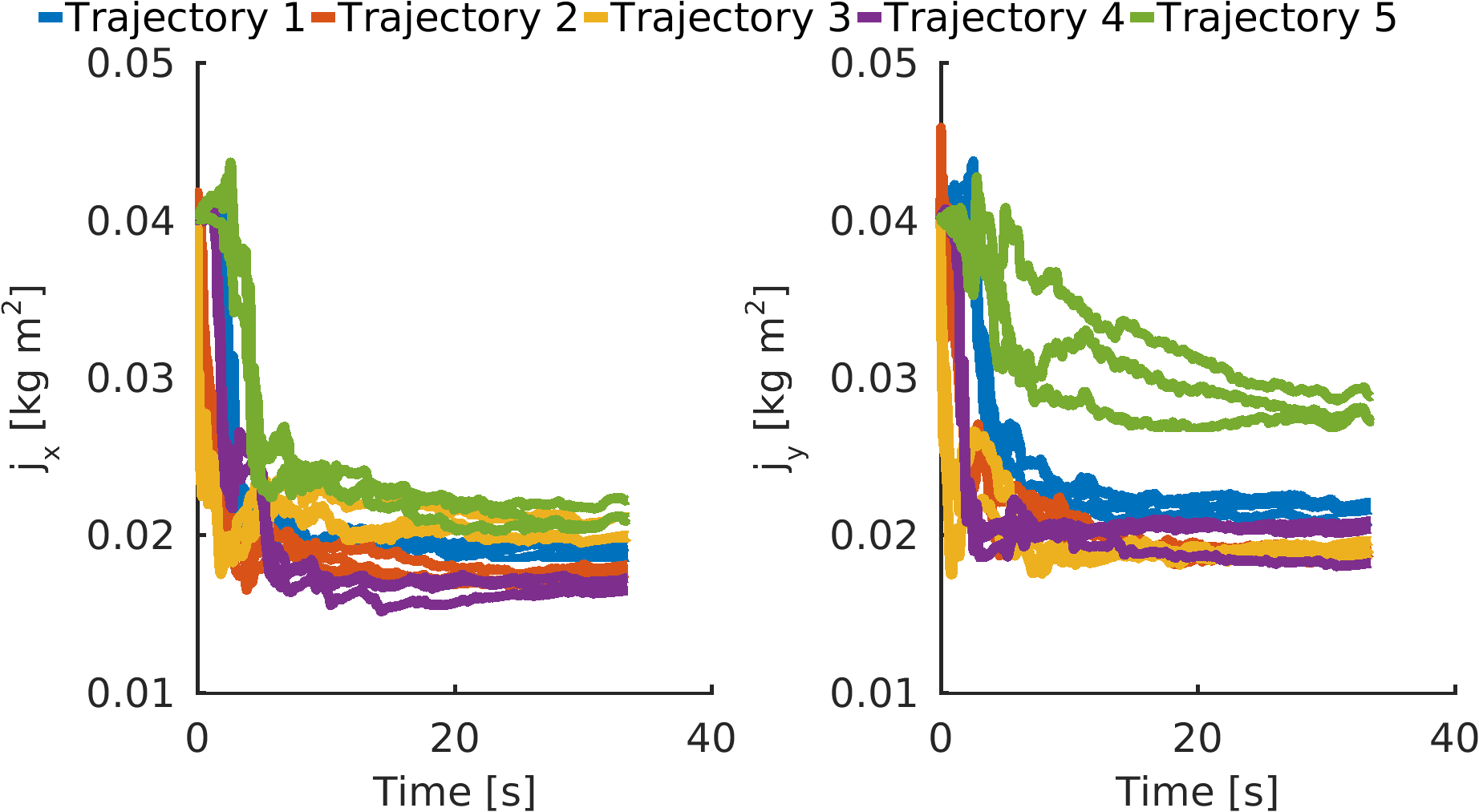}
\caption{$15$ real system identification experiments with $5$ different optimized trajectories.
Each trajectory is executed $3$ times to show the influence of the trajectory on the \ac{EKF} parameter estimation.
The values quickly converge towards reasonable values in less than $\SI{10}{\second}$.
Only for the green trajectory the value differs significantly.
We conclude that this trajectory most likely triggers effects that were not modeled in our \ac{EKF} implementation.}
\label{fig:real_est}
\vspace{-6mm}
\end{figure}

In a second set of experiments, we validated our approach on a real \ac{AscTec} Firefly with a Vicon motion tracking system and parameter independent nonlinear \ac{MPC} \cite{kamel2016linear}.
An example trajectory and the \ac{MAV} executing this trajectory are shown in \Figure{\ref{fig:eyecatcher}}.
\Figure{\ref{fig:real_est}} shows the actual parameter estimate outcome for $5$ optimized trajectories starting from a fixed initial guess.
Each trajectory was executed $3$ times to check the repeatability of the resulting estimates.
The planner excites roll and pitch such that $\inertiax$ and $\inertiay$ converge quickly, in less than \SI{10}{\second}, towards reasonable values.
We can observe that the estimates are consistent, but one of the trajectories, the green one, differs in the final value for $\inertiay$.

An important finding from this result is that our simplifications to the implemented model might be too strong for the real system.
Because the estimation bias occurs on the same trajectory repeatably, we conclude they stem from unmodeled aerodynamic effects and sensor delays.
The \ac{EKF} becomes overconfident on a particular motion.
We still think that this trajectory is informative and we suggest either using a different estimator, e.g., a batch optimization \cite{burri2016identification} to perform the actual estimation on the sensor data or a more elaborate \ac{MAV} model.

\section{Conclusion}
We have presented a motion planning algorithm for finding maximally informative system identification experiments.
The system is fully automated and therefore no expert knowledge about which motions to perform to achieve a good calibration is needed.
Furthermore, our planner requires minimum tuning effort and no teleoperated flight skills.
While we focus on \ac{MAV} model identification, one can easily adapt the framework for any kind of robot configuration.

Performing the sampling in the \acp{MAV} flat space and calculating a one-step propagation function between two samples, the planner is able to efficiently search the robot's belief space.
Unlike optimization based approaches our sampling-based approach can find a good solution after a few seconds, making it suitable for in-field calibration.
Our method also does not require any initial guess for the trajectory which is otherwise required to avoid falling into a local minimum.

We showed in simulation that, on average, in only \SI{30}{\second} runtime our planner creates trajectories which lead to more than $4\times$ quicker and more confident parameter estimates than random trajectories.
Real experiments confirm the feasibility of our approach and emphasize the importance of a good model and calibration routine for \ac{MAV} parameter estimation.

\balance
\bibliographystyle{bibliography/IEEEtran}
\bibliography{bibliography/IEEEabrv,bibliography/bibliography}

\begin{thebibliography}{10}
\providecommand{\url}[1]{#1}
\csname url@rmstyle\endcsname
\providecommand{\newblock}{\relax}
\providecommand{\bibinfo}[2]{#2}
\providecommand\BIBentrySTDinterwordspacing{\spaceskip=0pt\relax}
\providecommand\BIBentryALTinterwordstretchfactor{4}
\providecommand\BIBentryALTinterwordspacing{\spaceskip=\fontdimen2\font plus
\BIBentryALTinterwordstretchfactor\fontdimen3\font minus
  \fontdimen4\font\relax}
\providecommand\BIBforeignlanguage[2]{{%
\expandafter\ifx\csname l@#1\endcsname\relax
\typeout{** WARNING: IEEEtran.bst: No hyphenation pattern has been}%
\typeout{** loaded for the language `#1'. Using the pattern for}%
\typeout{** the default language instead.}%
\else
\language=\csname l@#1\endcsname
\fi
#2}}

\bibitem{kumar2012opportunities}
V.~Kumar and N.~Michael, ``Opportunities and challenges with autonomous micro
  aerial vehicles,'' in \emph{The International Journal of Robotics Research
  (IJR)}, vol.~31, 2012.

\bibitem{mina2015}
M.~Kamel, K.~Alexis, M.~Achtelik, and R.~Siegwart, ``{Fast nonlinear model
  predictive control for multicopter attitude tracking on SO(3)},'' in
  \emph{IEEE Conference on Control Applications (CCA)}, 2015.

\bibitem{burri2015robust}
M.~Burri, M.~Datwiler, M.~W. Achtelik, and R.~Siegwart, ``{Robust state
  estimation for Micro Aerial Vehicles based on system dynamics},'' in
  \emph{IEEE Conference on Robotics and Automation (ICRA)}, 2015.

\bibitem{Furrer2016}
F.~Furrer, M.~Burri, M.~Achtelik, and R.~Siegwart, \emph{{Robot Operating
  System (ROS): The Complete Reference (Volume 1)}}.\hskip 1em plus 0.5em minus
  0.4em\relax Springer International Publishing, 2016, ch. RotorS--A Modular
  Gazebo MAV Simulator Framework.

\bibitem{burri2016identification}
M.~Burri, J.~Nikolic, H.~Oleynikova, M.~{Achtelik W.}, and S.~Roland, ``Maximum
  likelihood parameter identification for mavs,'' in \emph{IEEE/RSJ
  International Conference on Intelligent Robots and Systems (IROS)}, 2016.

\bibitem{Hausman}
K.~Hausman, J.~Preiss, G.~S. Sukhatme, and S.~Weiss, ``{Observability-Aware
  Trajectory Optimization for Self-Calibration with Application to UAVs},''
  2016, arXiv:1604.07905.

\bibitem{Swevers1997}
J.~Swevers, C.~Ganseman, D.~B. T{\"u}kel, J.~De~Schutter, and H.~Van~Brussel,
  ``Optimal robot excitation and identification,'' in \emph{IEEE Transactions
  on Robotics and Automation}, vol.~13, no.~5, 1997.

\bibitem{Wilson2014}
A.~Wilson, J.~Schultz, and T.~Murphey, ``{Trajectory Synthesis for Fisher
  Information Maximization},'' in \emph{IEEE Transactions on Robotics (T-RO)},
  vol.~30, no.~6, 2014.

\bibitem{Wilson2015}
A.~D. Wilson, J.~A. Schultz, and T.~D. Murphey, ``{Trajectory Optimization for
  Well-Conditioned Parameter Estimation},'' in \emph{IEEE Transactions on
  Automation Science and Engineering (T-ASE)}, vol.~12, no.~1, 2015.

\bibitem{ko1995exact}
C.-W. Ko, J.~Lee, and M.~Queyranne, ``{An exact algorithm for maximum entropy
  sampling},'' in \emph{INFORMS Operations Research}, vol.~43, no.~4, 1995.

\bibitem{Hollinger2014}
G.~A. Hollinger and G.~S. Sukhatme, ``{Sampling-based Motion Planning for
  Robotic Information Gathering},'' in \emph{The International Journal of
  Robotics Research (IJR)}, vol.~33, no.~9, 2014.

\bibitem{Bry2011}
A.~Bry and N.~Roy, ``{Rapidly-exploring random belief trees for motion planning
  under uncertainty},'' in \emph{IEEE Conference on Robotics and Automation
  (ICRA)}, 2011.

\bibitem{Prentice2009}
S.~Prentice and N.~Roy, ``{The Belief Roadmap : Efficient Planning in Belief
  Space by Factoring the Covariance},'' in \emph{The International Journal of
  Robotics Research (IJR)}, vol.~28, 2009.

\bibitem{achtelik2014motion}
M.~W. Achtelik, S.~Lynen, S.~Weiss, M.~Chli, and R.~Siegwart, ``{Motion-and
  Uncertainty-aware Path Planning for Micro Aerial Vehicles},'' in
  \emph{Journal of Field Robotics (JFR)}, vol.~31, no.~4, 2014.

\bibitem{Carrillo2012}
H.~Carrillo, I.~Reid, and J.~A. Castellanos, ``{On the comparison of
  uncertainty criteria for active SLAM},'' in \emph{IEEE Conference on Robotics
  and Automation (ICRA)}, 2012.

\bibitem{mellinger2011minimum}
D.~Mellinger and V.~Kumar, ``{Minimum snap trajectory generation and control
  for quadrotors},'' in \emph{IEEE Conference on Robotics and Automation
  (ICRA)}, 2011.

\bibitem{richter2013polynomial}
C.~Richter, A.~Bry, and N.~Roy, ``{Polynomial trajectory planning for
  aggressive quadrotor flight in dense indoor environments},'' in
  \emph{International Symposium on Robotics Research (ISRR)}, 2013.

\bibitem{muellercomputationally}
M.~W. Mueller, M.~Hehn, and R.~D'Andrea, ``A computationally efficient motion
  primitive for quadrocopter trajectory generation,'' in \emph{IEEE
  Transactions on Robotics (T-RO)}, vol.~31, no.~6, 2015.

\bibitem{mahony2012multirotor}
R.~Mahony, V.~Kumar, and P.~Corke, ``{Multirotor aerial vehicles: Modeling,
  estimation, and control of quadrotor},'' in \emph{IEEE Robotics {\&}
  Automation Magazine}, vol.~19, no.~3, 2012.

\bibitem{weiss2012real}
S.~Weiss, M.~W. Achtelik, S.~Lynen, M.~Chli, and R.~Siegwart, ``Real-time
  onboard visual-inertial state estimation and self-calibration of mavs in
  unknown environments,'' in \emph{IEEE Conference on Robotics and Automation
  (ICRA)}, 2012.

\bibitem{weiss2012vision}
S.~M. Weiss, ``{Vision based navigation for micro helicopters},'' Ph.D.
  dissertation, ETH Z\"urich, 2012.

\bibitem{Trawny2005}
N.~Trawny and S.~I. Roumeliotis, ``{Indirect Kalman filter for 3D Attitude
  Estimation},'' in \emph{University of Minnesota, Department of Computing
  Science and Engineering, Tech. Rep.}, no. 2005-002, 2005.

\bibitem{omplbenchmark}
M.~Moll, I.~A. Sucan, and L.~E. Kavraki, ``{Benchmarking Motion Planning
  Algorithms: An Extensible Infrastructure for Analysis and Visualization},''
  in \emph{IEEE Robotics {\&} Automation Magazine}, vol.~22, no.~3, 2015.

\bibitem{ompl}
I.~A. Sucan, M.~Moll, and L.~E. Kavraki, ``{The Open Motion Planning
  Library},'' in \emph{IEEE Robotics {\&} Automation Magazine}, vol.~19, no.~4,
  2012.

\bibitem{kamel2016linear}
M.~Kamel, M.~Burri, and R.~Siegwart, ``Linear vs nonlinear mpc for trajectory
  tracking applied to rotary wing micro aerial vehicles,'' 2016,
  arXiv:1611.09240.

\end{thebibliography}

\newpage
\appendices

\end{document}